\documentclass[sigconf]{acmart}
\usepackage{graphicx}
\graphicspath{ {./figures/} }
\usepackage{algorithm}
\usepackage{algpseudocode}
\usepackage{algorithmicx}
\usepackage{multirow}
\usepackage{algorithm,algpseudocode}
\usepackage{xcolor}
\usepackage{todonotes}
\usepackage{xcolor}
\usepackage{soul}

\usepackage{amsmath}

\AtBeginDocument{%
  \providecommand\BibTeX{{%
    \normalfont B\kern-0.5em{\scshape i\kern-0.25em b}\kern-0.8em\TeX}}}

\renewcommand\footnotetextcopyrightpermission[1]{} 

\begin{document}

\title{Turning Privacy-preserving Mechanisms against Federated Learning}

\author{Marco Arazzi}

\affiliation{%
  \institution{University of Pavia}
  \streetaddress{}
  \city{Pavia}
  \state{}
  \country{Italy}
  \postcode{}
}
\email{marco.arazzi01@universitadipavia.it}

\author{Mauro Conti}
\affiliation{%
  \institution{University of Padua}
  \streetaddress{}
  \city{Padova}
  \country{Italy}}
\email{conti@math.unipd.it}

\author{Antonino Nocera}

\affiliation{%
  \institution{University of Pavia}
  \streetaddress{}
  \city{Pavia}
  \state{}
  \country{Italy}
  \postcode{}
  }
\email{antonino.nocera@unipv.it}

\author{ Stjepan Picek}

\affiliation{%
  \institution{Radboud University \& Delft University of Technology}
  \streetaddress{}
  \city{Nijmegen}
  \state{}
  \country{The Netherlands}}
\email{stjepan.picek@ru.nl}


\begin{abstract}
  Recently, researchers have successfully employed Graph Neural Networks (GNNs) to build enhanced recommender systems due to their capability to learn patterns from the interaction between involved entities.
  In addition, previous studies have investigated federated learning as the main solution to enable a native privacy-preserving mechanism for the construction of global GNN models without collecting sensitive data into a single computation unit.
  Still, privacy issues may arise as the analysis of local model updates produced by the federated clients can return information related to sensitive local data.
  For this reason, experts proposed solutions that combine federated learning with Differential Privacy strategies and community-driven approaches, which involve combining data from neighbor clients to make the individual local updates less dependent on local sensitive data.
  
  In this paper, we identify a crucial security flaw in such a configuration, and we design an attack capable of deceiving state-of-the-art defenses for federated learning.
  The proposed attack includes two operating modes, the first one focusing on convergence inhibition (\emph{Adversarial Mode}), and the second one aiming at building a deceptive rating injection on the global federated model (\emph{Backdoor Mode}).  
  The experimental results show the effectiveness of our attack in both its modes, returning on average $60\%$ performance detriment in all the tests on Adversarial Mode and fully effective backdoors in $93\%$ of cases for the tests performed on Backdoor Mode.
\end{abstract}



\maketitle
\thispagestyle{empty}
\pagestyle{plain}

\section{Introduction}
\label{sec:introduction}

In the last few years, federated learning has gained growing attention from the research community thanks to its capability of supporting privacy-preserving approaches for the construction of machine learning and deep learning models.
Indeed, with the massive availability of data that characterizes the current information technology realm, complex Artificial Intelligence (AI) solutions have been brought to life just by leveraging that data availability and the most recent technological advancements.
However, one of the baseline factors of the aforementioned data era is the diffused, user-level data production and collection.
On the one hand, this situation enables the design and realization of advanced AI products, but the user-level granularity of usable information has raised important privacy and security concerns on the other hand.
Moreover, in line with the technological advancements,  the legal aspects and regulations have received increasing attention~\cite{voigt2017eu}, imposing, in some cases, even a firm limit to AI progress, driving researchers to work on solutions where privacy protection becomes the main constraint.
This is precisely one of the objectives of federated learning, whose design allows for the training of deep learning models avoiding the need to centralize the collection of possible sensitive data into a single computation unit.
Indeed, according to this learning paradigm, the computation is distributed, and each involved client is responsible for the independent training of a local model using a private, non-shareable set of data.
A super node, which acts as an aggregation server, coordinates the training task by collecting, at each training epoch, the updates of the local models from the clients and by applying a suitable aggregation strategy to build the final global model.
Previous studies have proven the successfulness of his technology, mainly in its horizontal variant~\cite{yang2019federated}, applying it in many application contexts. 
This is especially true in scenarios in which social collaboration among users can provide important contributions to obtain improved sophisticated AI solutions.
For instance, very recently, researchers have adopted deep learning approaches in the context of recommender systems to refine the recommendation strategy.
In this context, they demonstrated that graph neural networks (GNNs) are very promising due to their ability to learn patterns from the interactions between the modeled entities~\cite{ying2018graph,wu2022graph}.
However, an open issue concerning the use of GNNs refers to the so-called cold-start, for which freshly involved users do not have enough interaction history so that the model can learn adequate representations of their profiles.
To overcome this limitation, some recent studies have leveraged additional social information of involved users~\cite{fan2019graph,mu2019graph,wu2019neural}.
In this way, the representation of a user can also be learned by analyzing his/her social neighborhood.
According to this strategy, the underlying GNN of such a recommender system will model both the classical user-item interactions and the social user-user ones.
Of course, the introduction of additional sensitive data related to the social life of users has urged even more researchers to rapidly define and adopt privacy-preserving strategies, with federated learning being the most prominent choice.

However, despite its distributed design representing a native privacy solution, researchers have shown that federated learning is vulnerable to attacks.
Thus, an adversary could infer sensitive information related to the original private data of local clients based on the local model variations recorded during consecutive learning epochs~\cite{lyu2020threats,chai2020secure,liu2022federated}.
To address this vulnerability, several recent studies have combined federated learning with Differential Privacy techniques~\cite{mcsherry2009differentially}.
This ensures that the rating assigned by users to items cannot be inferred by analyzing the local model updates of consecutive epochs. 
However, as stated above, to face the cold-start issue, also social information can be included, thus adding a higher complexity level to the whole solution.
Leveraging this information, some authors have proposed to exploit the social nature of the underlying scenario to create an additional collaborative privacy-preserving mechanism~\cite{liu2022federated,wu2021fedgnn}.
In practice, the idea underlying these strategies is to augment the training of the local models with information derived from the surrounding social neighborhood so that the produced updates will not be dependent only on the local data.
Interestingly, as shown in~\cite{liu2022federated}, such an augmentation mechanism not only addresses the privacy concerns discussed above but ultimately leads to the improved general performance of the global model. 

The proposal described in this paper starts from these recent research efforts. Our intuition is that although the additional social collaborative solutions can help both in improving the performance of considered systems and in building strong privacy-preserving approaches, this paradigm can be maliciously exploited to craft very impactful cyber attacks.
In general, the decentralized nature of federated learning makes it a very interesting target environment for attackers. Indeed, each involved client, as well as the aggregating server, can become potential adversaries to the system~\cite{fang2020local,bagdasaryan2020backdoor,yang2023model}.
For this reason, the research community has developed several countermeasures and advanced protection solutions that can be successfully exploited to protect this complex environment~\cite{nguyen2023poisoning,fung2020limitations,blanchard2017machine}.
However, by analyzing the behavior of the most recent defenses, we can see that the main strategy adopted therein is basically to detect and isolate from the system any action that differs from the average behavior of the community composing the federated scenario.
By contrast, the collaborative strategy introduced by the novel privacy-preserving mechanisms tries to protect the local contributions of single clients.
To do so, these strategies suitably combine local updates with those of the surrounding community members.
From an attacker's point of view, this configuration can become an opportunity to spread the attack to its neighbors. 
Thus, obtaining a novel threat possibly capable of even deceiving the state-of-the-art protections.

In this paper, we leverage this intuition to design a novel AI-based attack strategy for a scenario characterized by a social recommender system equipped with the privacy protection measures introduced above.
Borrowing some ideas from the related literature~\cite{baruch2019little}, we include two modes for the attack in our design, namely: a convergence inhibition strategy ({\em Adversarial Mode}) and a deceptive rating injection solution ({\em Backdoor Mode}).
More precisely, we implemented our proposal by focusing on the system described in~\cite{liu2022federated}, in which a GNN model is trained with a federated learning approach to build a social recommender system.
To achieve a strong privacy protection level, the target system includes both a Local Differential Privacy module and a community-based mechanism, according to which pseudo-items derived from the community are included in the local model training.
We argue that, although the attack described in this paper is specifically tailored to the features of such a system, the underlying intuition and methodology can be generalized to other similar scenarios.
The contributions of this paper can be summarized as follows:
\begin{itemize}
    \item We identify the main vulnerabilities of community-based privacy protection mechanisms for federated learning, focusing on approaches targeting Graph Neural Networks as underlying deep learning models.
    \item To deceive state-of-the-art security solutions for federated learning, we propose a model poisoning attack leveraging the features of the considered scenario.
    \item We adapt our attack to work in two modes: {\em Adversarial Mode} aiming at inhibiting the convergence of the federated learning model, and {\em Backdoor Mode} focusing on the creation of a backdoor in the learned model.
    \item To assess the performance of our attack, we adopt the Root Mean Squared Error, the Mean Absolute Error, and a newly defined metric, called {\em Favorable Case Rate} specific to estimate the success rate of our backdoor attack against the regressor that powers the recommender system.
    \item We test the effectiveness of our attack against a real-life recommender system based on the approach of~\cite{liu2022federated}.
    Moreover, we carried out an experimental campaign leveraging three very popular datasets for recommender systems.
    The obtained results show that our attack can cause very strong effects in both operating modes. In particular, in {\em Adversarial Mode}, it is capable of causing a $60 \%$ detriment in the performance of the target GNN model, on average, whereas, in {\em Backdoor Mode}, it allows the construction of fully effective backdoors in about $93\%$ of cases, also in the presence of the most recent federated learning defenses.
\end{itemize}

The remainder of this paper is organized as follows. In Section~\ref{sec:Background}, we describe some background concepts related to our reference scenario. Section~\ref{sec:SysteModel} describes the system model and the intuition underlying our attack.  
The technical details of our attack are discussed in Section~\ref{sec:Attack}. In Section~\ref{sec:Evaluation}, we report the experiments carried out to assess the effectiveness of our attack.
The related literature is surveyed in Section~\ref{sec:RW}. 
Finally, in Section~\ref{sec:Conclusion}, we draw conclusions and discuss potential future work directions.

\section{Background}
\label{sec:Background}

This section is devoted to the description of background concepts for our study.
In particular, we begin by introducing existing federated learning solutions that focus on privacy-preserving applications, with particular emphasis on recommender systems based on Graph Neural Networks.
After that, we describe model poisoning attacks in this context and introduce the most popular and effective countermeasures.

\subsection{Privacy-preserving Federated Learning}

Federated learning exploits decentralized parties, which own private sets of data to build global models through the suitable aggregation of learning information derived from the local training of individual models. This infrastructure ensures the construction of global models without sharing data between the involved parties. In any case, this scenario opens possible threats to the privacy of involved actors, including the possibility of inferring the private original data based on the model updates during the training phase or by observing the output produced subsequently.

In this context, solutions like Local Differential Privacy (LDP)~\cite{dwork2008differential} allow basic protection of the privacy of the federated learning clients by limiting the influence of the single datasets.
Generally speaking, Local Differential Privacy achieves privacy protection by norm clipping and adding noise to the updates of the local models from the clients. Some effective solutions apply Local Differential Privacy by adding Gaussian or Laplacian noise~\cite{liu2022federated}:
\begin{equation}
\Tilde{g}^c = clip(g^c, \gamma) + Laplacian(0, \lambda),
\end{equation}

\noindent
where $g^n$ are the updates of a client $c \in C$, $C$ is the set of clients, $\gamma$ is the clipping limit, and $\lambda$ is the standard deviation of the Laplacian noise.
As an example, in the approach of~\cite{sajadmanesh2021locally}, a Graph Convolutional Neural Network is trained in a federated way, and the privacy of the clients is preserved by using a Local Differential Privacy solution. Specifically, the involved clients protect their real updates from a potentially malicious data aggregator by providing a perturbed version of their updates that is not meaningful individually, which, however, guarantees the same training capability as the real ones when aggregated with the other contributions.
In addition, they also proposed a simple but effective Graph Convolutional Layer called $K-Prop$.
This layer aggregates messages from an extended neighborhood set,  which includes neighbor nodes with a distance of $K$ hops at maximum.
In this way, the proposed solution not only enhances client privacy by adding noise derived from real data but also improves the robustness of the global model because it is trained on an augmented dataset.

\subsection{Graph Neural Networks-based Recommender Systems}

By introducing links between users, social recommender systems compensate for the data sparsity problem. 
As typically done in Social Network Analysis, a very promising strategy in this setting is to model data through graphs, and then, ad-hoc Deep Learning algorithms, such as Graph Neural Networks, can be adapted to identify complex recommendation patterns.
Practically speaking, Graph Neural Networks are used to learn user and item embeddings from the graph to predict additional links between them. 
Works like the one proposed by Fan et al.~\cite{fan2019graph} exploit Graph Neural Networks, particularly Graph Attention Networks, to learn the embeddings of users and items for recommendation purposes.
In particular, this paper showed how using a GNN as an underlying model for a recommender system can be effective and efficient. The advantage of such models is the ability to aggregate high-order structural information that is important for learning embeddings from users and items.
Of course, due to the sensitivity of involved training data, this type of solution could also be implemented through a federated learning approach, in which data concerning the links between users and items remain locally private.
For instance, Wu et al.~\cite{wu2021fedgnn} proposed a federated learning approach to build a recommender system based on a GNN model collectively trained with highly decentralized user data.
This solution builds a robust model while preserving the privacy of the involved parties via Local Differential Privacy and user graph expansion, obtained by randomly sampling items from the neighbors. 

\subsection{Model Poisoning on Federated Learning}
\label{sub:backgroundPoisoning}

Due to its decentralized nature, federated learning introduces important security issues in scenarios where the involved clients cannot be assumed to be honest. In such a case, local model updates can be orchestrated by attackers to cause a detriment in the global model performance or, even worse, to drive the model behavior maliciously.
As described in Section~\ref{sub:Backgrounddefences}, to overcome these flaws, the aggregator entity of the federated learning solution can apply different robust aggregation strategies to limit the impact of such attacks.
These defense methods are, typically, Byzantine-robust algorithms that filter possibly malicious updates returned by the clients using statistical approaches.

For instance, a baseline strategy could be to exclude gradient updates too distant from the mean (outside the interval confidence) of the distribution of the updates of all the clients.
However, the recent scientific literature has demonstrated that these methods are still vulnerable to model poisoning attacks.

In this setting, Baruch et al.~\cite{baruch2019little} proposed one of the most well-known attacks trying to circumvent these defense strategies.
There, the authors defined two attack variations, namely {\em Convergence Prevention} and {\em Backdooring}.
In the first version of the attack, the attacker controls a small set of clients and tries to perturb their updates, within a statistically admissible range, with the objective of preventing the convergence of the model.
Gradients are perturbed by finding a deviation range from the mean that cannot be detected by defense methods based on statistical heuristics.
Specifically, the attack identifies the updates from local models with the maximum distance from the mean of the update distribution. Then it boosts this edge signal by replicating it in all the updates sent by the attacked clients.
Instead, the second attack they proposed is a {\em backdoor} attack in which the attacker poisons the model during the training phase to force the prediction of a specific target class against a controlled input pattern.
In practice, the attacker seeks a range of parameters that, if attacked, force the model to produce the desired label. A successful configuration must not affect the model's performance on benign inputs.
In our paper, we follow a similar strategy and design two different variants of our attack. 
In particular, our attack leverages the vulnerabilities introduced by the recent privacy-preserving techniques for GNN-based recommender systems trained through federated learning. As we show in our experiments, our attack proved to be more effective than the one presented in \cite{baruch2019little} also against the defense mechanisms described in the next section.

Still, in this context, Fang et al.~\cite{fang2020local} proposed another relevant example of a model poisoning attack.
In this case, the authors have defined two versions of the attack, the former referring to a situation in which the attacker has partial knowledge of the clients (i.e., the attacker knows only the controlled clients), and the latter, instead focusing on a condition in which the attacker has full knowledge of the federated learning scenario.
In both cases, the attacker crafts compromised local updates by maximizing or minimizing the parameters in such a way as to skew the global model in the reverse of the expected gradient direction; that is, the direction along which the global model would converge in a favorable situation.

\subsection{Defenses against Model Poisoning}
\label{sub:Backgrounddefences}

According to the basic implementation of a federated learning solution, the global model training is obtained by aggregating the local model updates returned by involved clients.
However, as explained above, this strategy introduces many security issues in general scenarios where the clients cannot be assumed fully secure.
Among the other security threats, model poisoning, either in the form of convergence prevention or backdooring, is, for sure, one of the most critical. Over the years, researchers have proposed several countermeasures for this reason.
In particular, Yin et al.~\cite{yin2018byzantine} proposed an enhanced version of the basic gradient aggregation strategy called {\tt TrimmedMean}. 
According to this solution, the server aggregates the gradients in the $i_{th}$ position independently.
Specifically, given the gradients of all the clients in the $i_{th}$ position, the aggregator sorts them according to their distance from the median.
Then, only the $top-k$ parameters are considered benign, where $k=n-m$, $n$ is the number of clients, and $m$ is the corrupted portion of them.

Blanchard et al.~\cite{blanchard2017machine} proposed a solution called {\tt Krum} that updates the global model by choosing the best candidates between the gradients returned by the clients.
The chosen gradients are those returned by the clients whose updates are the closest to the group of $n-m-2$ presumably honest workers.
The main intuition behind this approach is that, even if the selected updates are from malicious clients, they would still be close to the group of honest clients. According to this mechanism, all the outliers that differ significantly from the average will be discarded.
Both {\tt TrimmedMean} and {\tt Krum} are designed to work in a scenario with up to $m = (\frac{n}{2} + 1)$ malicious clients.

Recently, Nguyen et al.~\cite{nguyen2022flame} proposed an advanced defense mechanism for {\em backdoor} attacks, named {\tt FLAME}, which combines a clustering algorithm with an adaptive differential privacy strategy.
The workflow of {\tt FLAME} consists of three main steps, namely: filtering, clipping, and noising.
The objective of the first step is to filter malicious clients and select only those with the highest probability of being honest.
To do so, the authors perform a clustering over the pairwise cosine similarity distances among the updates received from the clients using HDBSCAN.
Specifically, they configured it to return a cluster that includes at least 50\% of the batch of clients. With this setting, the candidate cluster will contain the majority of clients, and all the remaining updates, possibly poisoned, are marked as outliers.
The second and third steps are dedicated to an adaptive differential privacy approach that estimates an effective clipping bound and a sufficient level of noise, such that the effect of the backdoor attack is removed while preserving the original performance of the model.
The clipping bound should be dynamically adapted to the decreasing trend of the gradients' $L_{2}-norm$.
It is performed by scaling the updates of the clients so that the $L_{2}-norm$ of the updates becomes smaller or equal to the chosen threshold.
The clipped updates are then aggregated to obtain the new global model.
The third step adds a certain amount of noise to the aggregated updates. This amount is determined by estimating a sensitivity value based on the distance between the clients' updates. In this way, the proposed strategy can override the contribution of the attack on the global model.

Recently, Fung et al.~\cite{fung2020limitations} proposed another defense solution with the name of {\tt FoolsGold}.
In any iteration, {\tt FoolsGold} adapts the learning rate of each client based on the similarity distance of the updates, also considering information derived from past iterations.
To measure the distance between the updates, as done by {\tt FLAME}, the cosine similarity is used.
Poisoning attacks usually affect specific features of the model, which can be identified by measuring the magnitude of model parameters in the output layer of the global model. Hence, the malicious updates can be removed or re-weighted.
Another key point of {\tt FoolsGold} is the exploitation of the history of the previous updates.
Indeed, as stated above, the similarity distance among the updates is computed by considering the current values returned by the clients and the values of the historical updates produced in the previous iterations.
This additional feature allows more accurate identification of malicious attempts to corrupt the federated learning task.

\section{System Model and Attack Intuition}
\label{sec:SysteModel}

This section is devoted to describing the reference scenario of our attack.
In particular, in Section~\ref{sub:FeSoG}, we present the essential concepts and definitions necessary to understand the scenario.
In Section~\ref{sub:DesignGoal}, we describe the main characteristics that introduce important advantages to the referring scenario but, at the same time, can be exploited by an attacker to perform an even more powerful exploit. 

\subsection{The System Model}
\label{sub:FeSoG}

\begin{figure*}[ht]
    \centering
    \includegraphics[scale=0.38]{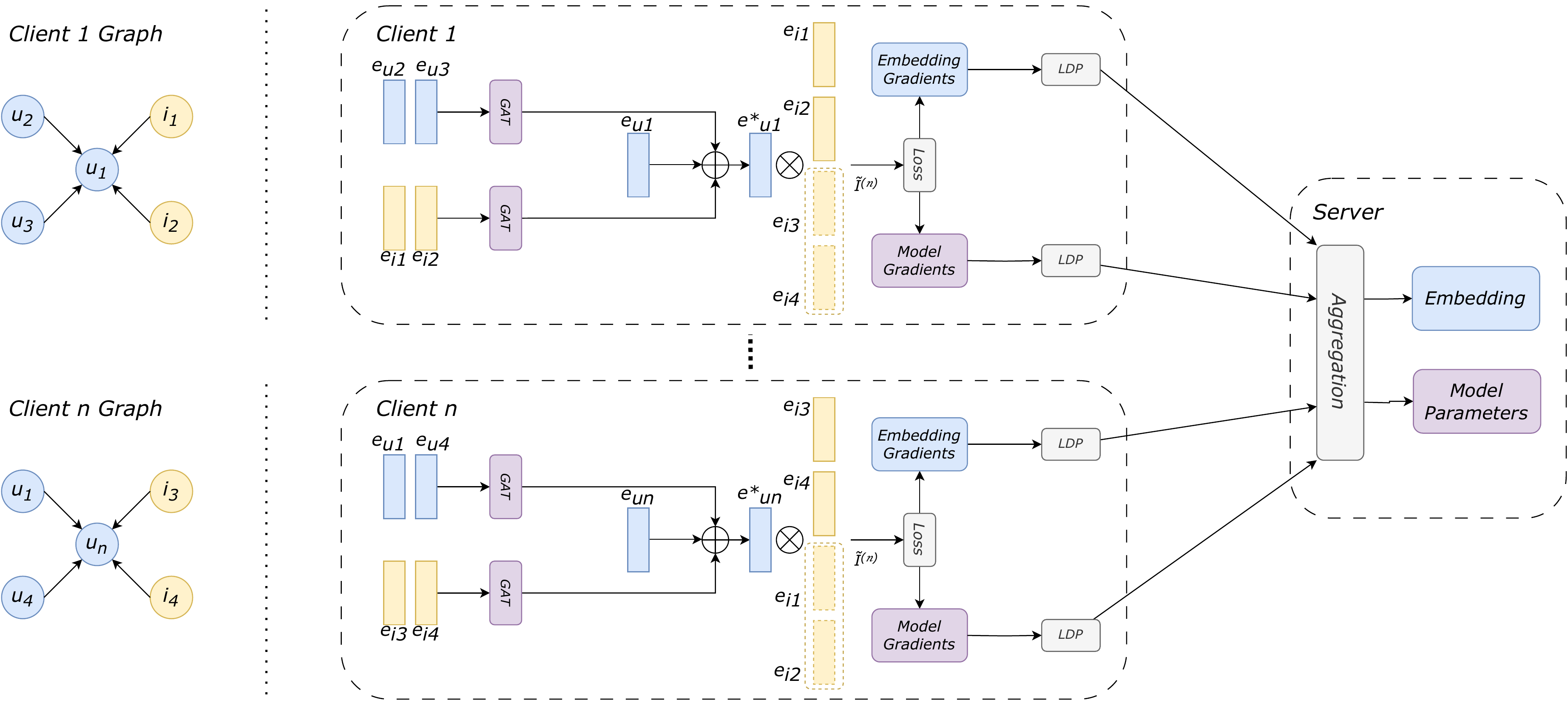}
    \caption{Main Scenario. \label{fig:MainScenario}}
\end{figure*}

The scenario for our approach is a privacy-aware social recommender system built through a federated learning solution.
To make our strategy concrete, we focus on a recent solution in this setting proposed in~\cite{liu2022federated}. 
It is worth observing that, although, in our approach, we make explicit reference to such a scenario, the main feature we are focusing on is a common strategy of social systems. Indeed, in such contexts, collaboration is generally leveraged to obtain joint advantages among peers.
Our strategy relies just on the fact that, if the common objective of the social system is to achieve privacy protection, such collaboration is typically ``blind'', and, even better, includes a Local Differential Privacy strategy, in such a way as to ensure non-disclosure of sensitive information.
We argue that, if properly handled, this condition can be exploitable to craft critical security menaces for social scenarios.

With that said, our target scenario, proposed by Liu et al.~\cite{liu2022federated} with the name {\tt FeSoG}, shown in Figure~\ref{fig:MainScenario}, is a federated social recommendation system (FSRS) designed to predict users' ratings for items using a Graph Neural Network model.
In this scenario, let $U = \{u_1, \dots, u_n\}$ be the set of users and $I = \{i_1, \dots, i_m\}$ be the set of items, where $N=|U|$ and $M=|I|$ are the number of users and items, respectively.
{\tt FeSoG} is composed of a set of clients $C = {c_1, \dots, c_n}$ such that each client $c_n$ is associated with a user $u_n$.
Due to this direct association, in the following, we shall use the terms user and client interchangeably.

The coordination of the federated training is delegated to a central unit, which receives the updated gradients from the clients and builds a global model by suitably aggregating them. 
By design, each client owns a local graph that contains the first-order neighbors and the information about the items of interest for the corresponding user along with their ratings.
Therefore, the local graph $G_n$ of a client $c_n$ consists of both user nodes and item nodes. 
$G_n$ is characterized by two types of edge, namely the {\em user-item} weighted edges, in which the weights represent the ratings assigned to the items by the corresponding users, and the {\em user-user} edges denoting the interactions between users. 

For each client, the set of rated items is denoted as $I^{(n)} = \{i_1, \dots, i_z\}$, whereas the set of neighbors is denoted as $U^{(n)} = \{u_1, \dots, u_k \}$.
Users and items are associated with their embeddings respectively $E_{u_n} \in \mathbb{R}^{dxN}$ and $E_{i_m} \in \mathbb{R}^{dxN}$, where $d$ is the dimension size of the embeddings.
A complete embedding table is maintained by the server and the clients can request access to this table.

By downloading the complete embedding table, a client can access the embeddings of the users and items that are part of its local graph $G_n$.
Such embeddings are, then, used as input for the local GNN model and, in particular, a GAT layer, to learn the embedding of the user associated with the client and predict the item scores.

In particular, the client embedding is an aggregation of both the embeddings of its neighbors and the embeddings of the rated items.
At this point, to predict the local item ratings for a specific user, the authors adopt a dot-product between the inferred user embedding and the item embeddings:
$$\hat{R}_{u_n,i_m}=E_{u_n} \cdot E_{i_m}.$$

One of the specifics of {\tt FeSoG} is the particular attention given to the privacy of the produced embeddings.
In particular, two techniques are implemented to protect the updates of the local user-item gradients, namely: Local Differential Privacy and pseudo-item labeling. 
The Local Differential Privacy solution prevents the user's rating information to be inferred, given the gradients uploaded by a user during two consecutive steps. 
To protect the gradients, each client clips its updates based on their $L_{2}-norm$ with a threshold $\gamma$ and adds a zero-mean Laplacian noise to achieve privacy protection.
The local differential privacy process is applied to the item embedding gradients $g_i^{(n)}$, the user embedding gradients $g_u^{(n)}$, and the model gradients $g_m^{(n)}$ for each client $c_n$.
This process can be formalized as follows:

 \begin{equation}
\Tilde{g}^{(n)} = clip(g^{(n)}, \gamma) + Laplacian(0, \lambda \cdot mean(g^{(n)})),
\end{equation}

\noindent
where $g^{(n)} = \{g_i^{(n)}, g_u^{(n)},  g_m^{(n)}\}$ is the combination of the gradients of the three different embeddings considered above.
Observe that because the involved gradients can be of a different magnitude, instead of applying a constant noise with strength $\lambda$, in this scenario, a dynamic noise is applied by multiplying $\lambda$ by the mean of the gradients themselves.

The second privacy-preserving technique introduced in this approach, instead, consists of the inclusion of pseudo-items in the training process of each local model. This guarantees an enhancement of user privacy, and, at the same time, an improvement of the robustness of the aggregated global model.
In practice, before the computation of the training loss on the local model, each client samples $p$ items $\Tilde{I}^{(n)} = \{\Tilde{i}_1^{(n)} \dots \Tilde{i}_p^{(n)}\}$, not already included in their local items.
Of course, for these additional pseudo-items only the corresponding embeddings are available to the client (through the embedding table available from the server). As for the corresponding ratings, a semi-supervised strategy is adopted, according to which the client uses its current local model to predict them for each pseudo-items.
At this point, such pseudo-items are included in the local loss computation as follows:
\begin{equation}
\label{eq:loss}
    L_{u_n} = \sqrt{\frac{\sum_{i_m \in I^{(n)} \cup \Tilde{I}^{(n)}} (R_{u_n,i_m} - \hat{R}_{u_n,i_m})}{ \left |I^{(n)} \cup \Tilde{I}^{(n)} \right |}},
\end{equation}

\noindent
where the adopted loss is the Root Mean Squared Error between the predicted ratings $\hat{R}_{ui}$ and the ground-truth rating scores $R_{ui}$. The pseudo-item sampling provides additional rating information, similar to data augmentation, which, in addition to improving the protection against data leakage, enhances the robustness of the local model.

\subsection{Attack Intuition and Challenges}
\label{sub:DesignGoal}

By design, the referring scenario introduces two main techniques that aim to improve the privacy protection of clients' data, while ensuring greater robustness of the global model built through federated learning.
Among them, the design choice of including pseudo-items in the local embeddings of clients plays a crucial role.
Indeed, as stated in Section~\ref{sub:FeSoG}, because the adopted pseudo-items are generated from real-data embeddings gathered from other clients, the introduced noise is informative and resembles a data augmentation solution.
On the other hand, in the case in which the assumption of the trustworthiness of clients does not hold, such an approach could lead to exploit opportunities for attackers.

The goal of this paper is to demonstrate that by leveraging this privacy-preserving social collaboration mechanism, it is possible to design a very powerful poisoning attack.
As will be shown in the experiments described in this paper, the social nature of such an attack allows the achievement of considerable performance also in the presence of cutting-edge defense solutions.

More in detail, the social mechanism of sampling pseudo-elements of peer clients to improve privacy protection, allows the possibility of involving such peers in prearranged attacks and, therefore, forcing them to include poisoned elements in their local training process, unknowingly.
Our attack aims, therefore, at performing a model poisoning by forging a malicious set of item embeddings. Our objective is to deceive the target recommender system and make it act as intended by the attacker, either by inhibiting convergence of the underlying GNN model or by performing a {\em backdoor} attack to force the system to predict specific ratings for items in relation to a target user.
In such a context, not only pseudo-items could be exploited, but also the Local Differential Privacy strategy could play a key role in the attack process.
As a matter of fact, many countermeasures, like for example the one proposed by Nguyen et al.~\cite{nguyen2022flame}, make use of Differential Privacy to override or erase the contribution of an attack, thus filtering malicious gradients updates in a federated learning solution. 
In our case, the Local Differential Privacy module, which is included in the privacy-aware social recommender system, acts as a regulator of the attack so that the poisoned changes to updates are the most similar to benign ones, while still guaranteeing the effectiveness of the attack.

\section{Attack Description}
\label{sec:Attack}

\begin{figure*}[ht]
    \centering
    \includegraphics[scale=0.4]{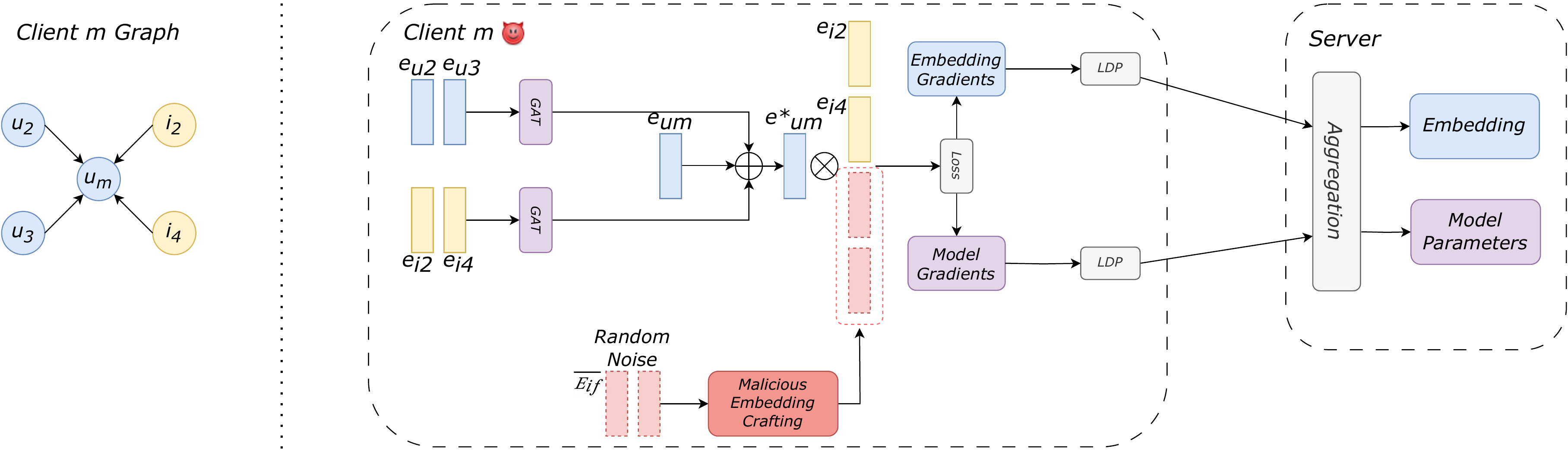}
    \caption{Attack Scenario. \label{fig:AttackScenario}}
\end{figure*}

This section is devoted to the design of an attack strategy against the target scenario introduced in the previous section, a schematic representation of which is shown in Figure~\ref{fig:AttackScenario}.

Similarly to the work proposed by Baruch et al.~\cite{baruch2019little}, our design includes two attack modes.
The former aims at the convergence of the aggregated model and attempts to significantly reduce its general performance.
The latter, instead, focuses on a more refined model poisoning goal, which is the construction of a {\em backdoor}.
In practice, it aims at forcing the model to predict specific ratings for items in relation to a target user.
Both attacks try to exploit vulnerabilities exposed by the strategy adopted to enhance the privacy and the robustness of the federated learning model, as described in detail in Section~\ref{sub:FeSoG}.

In the next sections, we shall report all the details related to the two attack types mentioned above. In particular, the former is presented in Section~\ref{sub:ConvergencePrevention}, and the latter is described in Section~\ref{sub:Backdoor}.

\subsection{Adversarial Mode - Convergence Inhibition}
\label{sub:ConvergencePrevention}

As presented in Section~\ref{sub:FeSoG}, according to the target scenario, each client involved in the privacy-aware social recommender system can sample a set of $p$ items, namely $\Tilde{I}^{(n)}$, from the pool of other clients in their neighborhood (according to the graph underlying the GNN), and assign them a pseudo-label.
This strategy allows them to add an {\em informative} noise to their local updates, thus producing two important effects: a higher privacy protection level and improved robustness of the final model.

The intuition behind our attack is that an attacker can exploit such a community-driven privacy-preserving mechanism, based on the sampled item set $\Tilde{I}^{(n)}$, to poison the federated learning model. 
We assume that the adversary can control a set, even small, of clients, hereafter referred to as malicious clients.
We argue that, by suitably crafting a poisoned item set, say $\overline{I}^{(n)}$, it might be possible to coerce the community around a malicious node to unwittingly participate in the attack, thus producing a hardly-detectable community attack. 

\begin{figure}[ht]
    \centering
    \includegraphics[width=0.85\columnwidth]{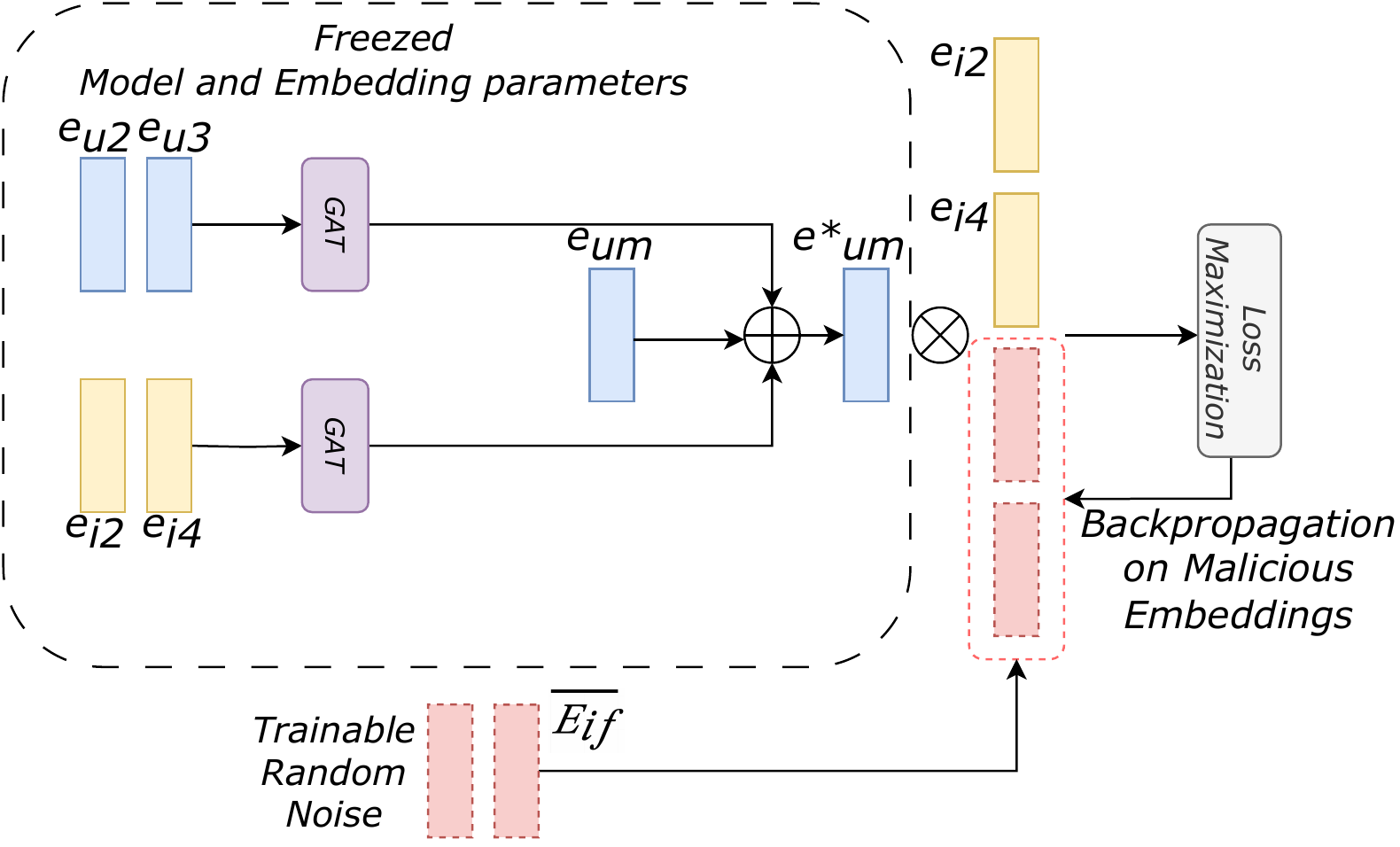}
    \caption{A schematic view of the proposed Convergence Prevention attack. \label{fig:ConvergencePrevention}}
\end{figure}

To do so, instead of sampling the items randomly from the other users, a malicious client tries to generate a set of fake embeddings $\overline{E_{i_f}} \in \mathbb(R)^{dxN}$ having the same shape $dxN$ obtained by sampling real items in normal conditions and, hence, corresponding to an implicit set of fake pseudo-items $\overline{I}^{(n)}$.
In particular, to undermine the convergence of the federated learning model, according to our strategy, starting from random Gaussian noise, at each training epoch $t$, the attacker trains malicious embeddings $\overline{E_{i_f}}$ to maximize the loss of the global model.
For this purpose, it uses the model parameters obtained from the server after the previous epoch $t-1$. Then, it performs a gradient descent optimization on the local model by keeping all the parameters frozen, with the exception of malicious embeddings.
In practice, to obtain effective malicious embeddings, an attacked client $c_n$ associated with a user $u_n$ pursues the following objective:
$$
\min_{\overline{I}^{(n)}} \left (- \sqrt{\frac{\sum_{i_m \in I^{(n)} \cup \overline{I}^{(n)}} (R_{u_n,i_m} - \hat{R}_{u_n,i_m})}{ \left |I^{(n)} \cup \overline{I}^{(n)} \right |}} \right ),
$$

\noindent
where, once again, the ratings of the fake pseudo-items are derived through a semi-supervised approach using the version of the local model obtained after epoch $t-1$.
Figure~\ref{fig:ConvergencePrevention} shows a graphical representation of the strategy above.

Once the malicious fake pseudo-items have been crafted, the attacker trains the local model, as done by any other client in the scenario, using the crafted embeddings $\overline{E_{i_f}}$ instead of the real embeddings $E_i$ of the sampled items $\Tilde{I}^{(n)}$.
It is worth observing that a domino effect is triggered by this strategy. Indeed, in doing so, the attacker poisons not only the updates of the local model but also the embeddings of the corresponding user, its neighbors, and the associated items. Moreover, the pseudo-item sampling task of the subsequent training epoch ($t+1$) of the federated learning will also include the current malicious embeddings introduced by the attacker. This will boost the exploit even more by involving other clients as unaware but still effective attackers.

\subsection{Backdoor Mode - Deceptive Rating Injection}
\label{sub:Backdoor}

The objective of the second attack mode is to poison the federated learning model in such a way that, given a target user $u_t$ and a set of target items $I^{(n)}_x$ not belonging to the local item set of $u_t$, a backdoor is created on the prediction of the ratings. 
In practice, the attacker aims to perform a backdoor attack that will force the recommender system to predict for the target user a specific (false) rating for these items. Thus, the adversary can even force the recommender system to always or never propose a specific item to a user based on the rating predicted by the model.
To carry out this attack, all the malicious clients controlled by the attacker must agree on the same target user $u_t$, a set of items $I^{(n)}_x$, and the target fake ratings to associate with them as a result of the poisoning action.
As for the considered scenario, the high-level objective of the attacker might be to force the inclusion (resp. exclusion) of the target items in the recommendation set.

To do so, instead of sampling a random set of pseudo-items $\Tilde{I}^{(n)}$, all the malicious clients use the same target set of pseudo-items $I^{(n)}_x$ and include it in the training of their local models.
As presented in Section~\ref{sub:FeSoG}, in our scenario, each client corresponding to a single user concatenates the embeddings of the local items (the items related to the underlying user) to the embeddings of the pseudo-items (sampled from the items related to the neighbor users) and, then, computes the corresponding ratings by combining them (i.e., applying a dot-product) with the trained user embedding.
To perform our attack instead of concatenating the local items $I^{(n)}$ with a set of pseudo-items $\Tilde{I}^{(n)}$ sampled at random, a malicious client $c_n$ performs the following steps:

\begin{itemize}
    \item First, it combines the local items embeddings $I^{(n)}$ with the embedding of the underlying user $u_n$, to obtain a rating prediction for such items according to $u_n$.
    \item Then, it combines the embeddings of the target items $I^{(n)}_x$ with the embedding of the target user $u_t$ to obtain a rating prediction for the items of $I^{(n)}_x$ according to $u_t$.
    \item Finally, it concatenates all the predicted ratings (of both the local items and the target ones) and uses them to calculate the loss (see Eq.~\eqref{eq:loss}), which is, hence, modified as follows:
       
    $$L_{u_n} = \sqrt{\frac{\sum_{i_m \in I^{(n)}} (R_{u_n,i_m} - \hat{R}_{u_n,i_m}) + \sum_{i_f \in I^{(n)}_x} (R_{u_t,i_f} - \hat{R}_{u_t,i_f})}{ \left |I^{(n)} \cup I^{(n)}_x \right |}}.$$
    
\end{itemize}

\begin{figure}[ht]
    \centering
    \includegraphics[width=0.85\columnwidth]{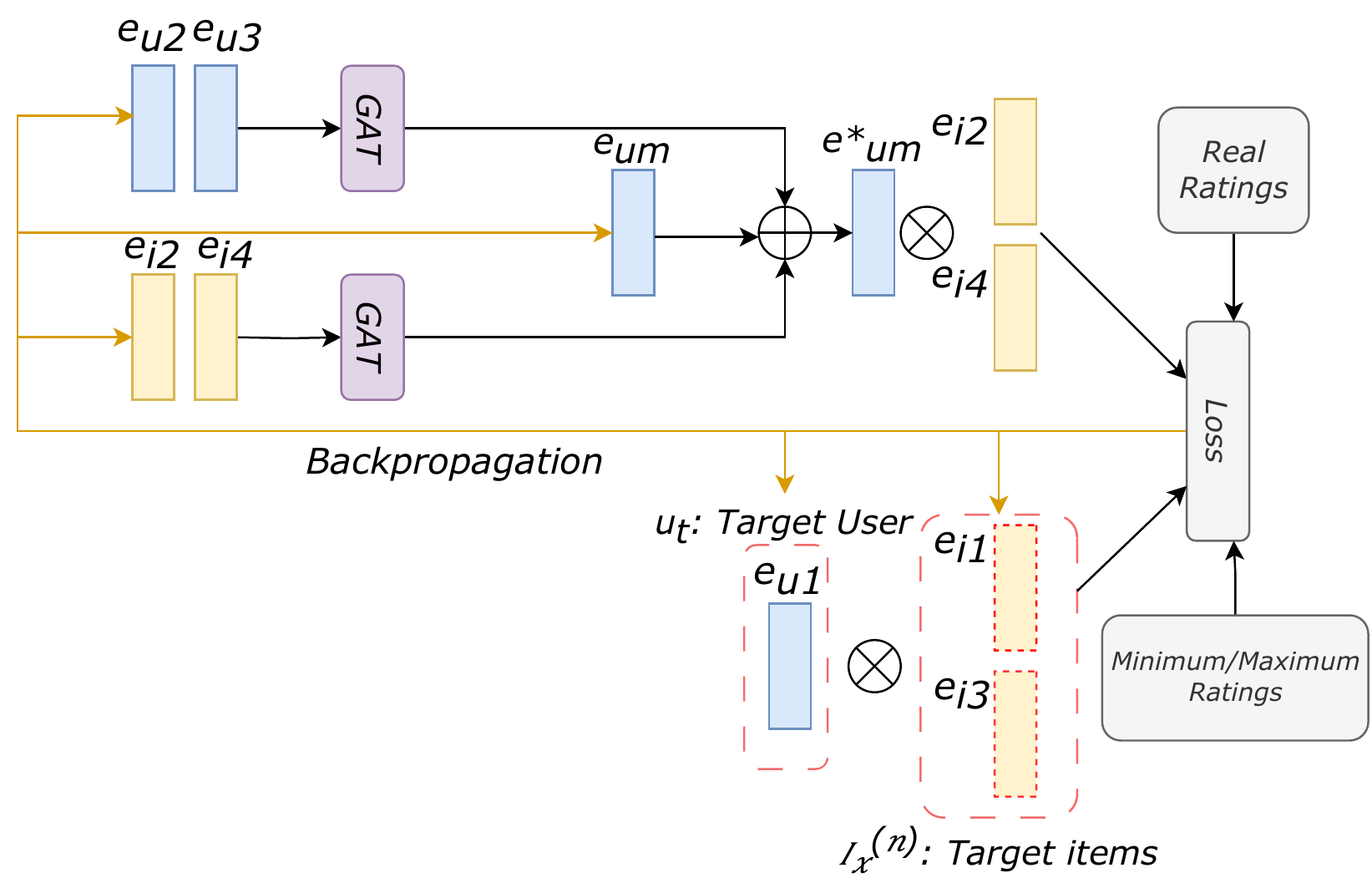}
    \caption{A schematic view of the proposed Backdoor attack. \label{fig:Backdoor}}
\end{figure}

As for the last point above, the value of the ground-truth rating score $R_{u_t,i_f}$ of Eq.~\eqref{eq:loss} for each target item, say $i_f$, of $I^{(n)}_x$, is forged by the attacker to obtain the desired effect on the final prediction (e.g., obtaining the maximum/minimum rate or setting it to a specific value).
In this way, the backpropagation on the model will include both the real signal from the local graphs of the clients controlled by the attacker and the additional poisoned knowledge designed to control only the rating scores for the items of $I^{(n)}_x$ for the target $u_t$.
Figure~\ref{fig:Backdoor} shows a representation of the steps described above.

\section{Attack Evaluation}
\label{sec:Evaluation}

In this section, we present the experiments carried out to assess the performance of both our attack modes on the referring scenario.
In particular, in Section~\ref{sub:Setup}, we describe the reference testbeds for our experiments. Sections~\ref{sub:ExpConvergence} and~\ref{sub:ExpBackdoor} are devoted to analyzing the results and performance of our attacks against different settings and defense mechanisms.

\subsection{The Considered Testbeds}
\label{sub:Setup}

To assess the performance of our attack, we define some reference testbeds, including the adopted evaluation metrics and the underlying datasets. Moreover, we identify the experimental setup by selecting the most promising configurations to properly test our solution.

\paragraph{\bf Evaluation Metrics.} To evaluate the effectiveness of our attack, we adopt the Mean Absolute Error and the Root Mean Squared Error and compare the performance of the target scenario in normal conditions and under our attack. For both metrics, smaller values are associated with better performance. For our {\em Convergence Prevention} attack, the exploit is successful when both the metrics above return higher values for the underlying GNN model (the deep model at the basis of the reference social recommender system) than in a condition with no attacks. 

As for the second attack type proposed in this paper, a successful backdoor must not affect the general performance of the target GNN model. Moreover, to further assess the effectiveness of the obtained backdoor, we define a metric called $Favarable \ Case \ Rate$ $(FCR)$.
As visible in Algorithm~\ref{alg:FCR}, this metric returns the percentage of target items whose residuals are lower than the Standard Error of the estimate function for good items. The objective of this metric is to assess whether the error produced by the model on the target items, with respect to the rating value aimed by the attacker, is comparable to the average baseline error rate obtained for good items (we require that this error is even lower than the average to declare an attack success). Indeed, such a condition would imply that the built backdoor successfully changes the behavior of the attacked model, forcing it to predict, for the target items, the ratings imposed by the attacker.

\begin{algorithm}[ht] 
\scriptsize
\caption{Favorable Cases Rate Function. \label{alg:FCR}}
\begin{algorithmic}[1]
\Require{\\ $Res$ (Residuals of real data) \\ $Res_{tr}$ (Residuals of the targeted ratings) \\ $SEE()$ (Standard error of estimate function)} 
\Ensure{ \\ $FCR$ (Favorable Cases Rate)}
\Statex
\Function{success\_rate}{$Res, \ Res_{tr}$}
  \State {$FCR$ $\gets$ {$0$}}
    \State {$N$ $\gets$ {$length(Res_{tr})$}}
    \For{$r$ in $Res_{tr}$}   
        \If{$r \ < \ SEE(Res)$}
            \State {$FCR$ $\gets$ {$FCR + 1$}}
        \EndIf
    \EndFor
    \State \Return {$FCR/N$}
\EndFunction
\end{algorithmic}
\end{algorithm}

\paragraph{\bf Datasets.} To validate our proposal, we adopt the same datasets used in~\cite{liu2022federated} to test the performance of the reference social recommender system (see Section~\ref{sub:FeSoG}). In particular, we use three  popular recommendation system datasets, namely: Ciao~\cite{tang2012mtrust}, Epinions~\cite{Tang-etal12b, Tang-etal12c, Tang-etal13a}, and Filmtrust~\cite{guo2013novel}. Ciao and Epinions have been collected by crawling shopping websites, and both of them are characterized by items rated with integers in the interval $(0,5)$ and social trust links among users. Similarly, Filmtrust is composed of a set of users connected by trust links and a set of items, each associated with a rating score ranging in the interval $(1,8)$. 

For our experiments, each user of the previous datasets is associated with a client, and the corresponding local graph is generated using the items that they have rated and the users with whom they have trust links to build their neighborhood. The statistics of the obtained datasets are reported in Table~\ref{tab:DatasetStats}.

\begin{table}[ht]
\caption{Statistics of the reference datasets \label{tab:DatasetStats}}
\scriptsize
\centering
\begin{tabular}{|l|c|c|c|}
\hline
Dataset & Ciao & Epinions & Filmtrust \\
\hline
Users        & 7,317 & 18,069 & 874\\
Items       & 104,975 &  261,246 &  1,957 \\
\hline
\# of ratings  & 283,320 & 762,938 & 18,662 \\
Rating density & 0.0369\% & 0.0162\% & 1.0911\% \\
\hline
\# of social connections & 111,781 &  355,530 & 1,853\\
Social connection density & 0.2088\% & 0.1089\% & 0.2426\% \\
\hline
\end{tabular}
\end{table}

\paragraph{\bf Experimental Setup.} The reference datasets are randomly split into three subsets: training set (60\%), validation test (20\%), and testing set (20\%). 
As for the validation set, it is used to evaluate the performance of the model during the training phase. 
In our configuration, the policy for the training early stopping, the learning rate, the initialization of the embeddings, and the strength of the Laplacian noise, are set as proposed in the reference scenario originally described in~\cite{liu2022federated}.
Specifically, the training process is stopped when the model does not improve on the validation set for more than $5$ successive validation steps. When the training phase is completed, the model is evaluated on the testing set. For the backdoor mode of our attack, at each validation step, we also assess the effectiveness of the attack on the target items. For all our experiments, the learning rate of the model is set to $0.01$, the embeddings are initialized with standard Gaussian distribution. 
Moreover, the gradient clipping threshold is set to $0.3$, and the strength of Laplacian noise is set to $0.1$. 
Finally, we tested our attack for different numbers of items sampled, specifically $\{10, \ 20, \ 30, \ 40, \ 50, \ 100\}$, and different percentages of attackers, namely, $\{10\%, \ 20\%, \ 30\%, \ 40\%, \ 50\%\}$.

\subsection{Results: Adversarial Mode}
\label{sub:ExpConvergence}

In this section, we analyze the performance of our attack in Convergence Prevention mode against the scenario introduced in Section~\ref{sub:FeSoG} ({\em Main Scenario}, for short). In our experiment, as an initial configuration, we set the percentage of attackers to $30\%$ of the total number of clients and the maximum number of pseudo-items sampled equal to $10$.
Moreover, as commonly done in this context, we also consider different protection configurations based on the most common and effective Federated Learning defenses, namely: {\tt Krum}, {\tt TrimmedMean}, {\tt FoolsGold}, and {\tt Flame} (see Section~\ref{sub:Backgrounddefences} for background on these defenses). 
Moreover, to provide a comparison baseline for the assessment of the effectiveness of our solution, we report: {\em (i)} the basic performance of the considered GNN model without the additional privacy-aware social mechanism proposed in~\cite{liu2022federated} based on pseudo-items ({\em Baseline Scenario}), {\em (ii)} the performance obtained in the same configuration when the system is attacked by a reference state-of-the-art attack, i.e., the Little Is Enough attack (LIE) (Section~\ref{sub:backgroundPoisoning}), {\em (iii)} the performance obtained by the complete solution of~\cite{liu2022federated} in the absence of attacks, and {\em (iv)} the performance obtained in the same configuration when the attack on the pseudo-items is performed using a naive strategy based on the generation of Gaussian noise.
The results on the three datasets introduced above are reported in Table~\ref{tab:ResConvergence}. 

\begin{table*}[ht]
\centering
\scriptsize
\caption{Results of the convergence inhibition attack\label{tab:ResConvergence}}
\begin{tabular}{|c|c|c|cc|cc|cc|}
\hline
\multirow{2}{*}{Scenario} & \multirow{2}{*}{Attack} & \multirow{2}{*}{Defense} & \multicolumn{2}{c|}{Filmtrust} & \multicolumn{2}{c|}{Ciao} & \multicolumn{2}{c|}{Epinions}\\ \cline{4-9} 
&  & & \multicolumn{1}{c|}{RMSE} & MAE & \multicolumn{1}{c|}{RMSE} & MAE & \multicolumn{1}{c|}{RMSE} & MAE\\ \hline
Baseline Scenario  & None & None & \multicolumn{1}{c|}{2.19}  & 1.60    & \multicolumn{1}{c|}{2.54}  & 1.87  & \multicolumn{1}{c|}{2.17} & 1.52  \\ \hline
Baseline Scenario  & LIE~\cite{baruch2019little} & None & \multicolumn{1}{c|}{2.37}    & 1.69    & \multicolumn{1}{c|}{2.80}          & 2.04  & \multicolumn{1}{c|}{2.36}          & 1.66\\ \hline 
Main Scenario & None & None & \multicolumn{1}{c|}{2.08} & 1.56    & \multicolumn{1}{c|}{2.18}          & 1.55 & \multicolumn{1}{c|}{1.79}          & 1.35\\ \hline
Main Scenario & Gaussian Noise & None & \multicolumn{1}{c|}{2.06}          & 1.57    & \multicolumn{1}{c|}{2.20}          & 1.59 & \multicolumn{1}{c|}{1.78}          & 1.36\\ \hline \hline

Main Scenario   & Our attack (Adversarial Mode) & FoolsGold & \multicolumn{1}{c|}{3.21}          & 2.69 & \multicolumn{1}{c|}{3.07}          & 2.45 & \multicolumn{1}{c|}{2.79}          & 2.51\\ \hline

Main Scenario   & Our attack (Adversarial Mode) & Flame & \multicolumn{1}{c|}{3.01}          & 2.30 & \multicolumn{1}{c|}{3.05}          & 2.45 & \multicolumn{1}{c|}{2.69}          & 2.34\\ \hline

Main Scenario   & Our attack (Adversarial Mode) & Krum & \multicolumn{1}{c|}{3.03}          & 2.44    & \multicolumn{1}{c|}{3.02}          & 2.42 & \multicolumn{1}{c|}{2.71}          & 2.35\\ \hline

Main Scenario  & Our attack (Adversarial Mode) & TrimmedMean & \multicolumn{1}{c|}{3.23}          & 2.60    & \multicolumn{1}{c|}{3.00}          & 2.42 & \multicolumn{1}{c|}{2.66}          & 2.31\\ \hline \hline
 \multicolumn{3}{|c|}{{\bf Average Performance Detriment}} & \multicolumn{1}{c|}{{\bf -50\%}}          & {\bf -60\%}    & \multicolumn{1}{c|}{{\bf -39\%}}          & {\bf -57\%} & \multicolumn{1}{c|}{{\bf -51\%}}          & {\bf -76\%}\\ \hline
\end{tabular}
\end{table*}

By analyzing this table, it is possible to see that our attack is capable of significantly decreasing the performance of the GNN model, with a performance reduction spanning from $39\%$ to $76\%$ with respect to the scenario in the absence of attacks. This result is even more astounding when we consider that, for the Baseline Scenario, the state-of-the-art LIE attack produces a maximum performance penalty of $10.2\%$. 
The obtained result also confirms that the use of community-derived pseudo-items and, in general, collaborative strategies to achieve privacy protection improves the robustness of the federated learning model (as originally shown in~\cite{liu2022federated}) but, at the same time, provides an adversary with the means to perform possibly stronger attack. As presented in Section~\ref{sub:ConvergencePrevention}, our attack crafts the embeddings of the pseudo-item by maximizing the loss of the model at each epoch. To assess the reasoning behind our strategy, in Table~\ref{tab:ResConvergence}, we report the results obtained by a basic attack in which, instead of learning optimal embeddings at each iteration, they are initialized with Gaussian noise. As we can clearly see from this table, the attack on pseudo-items using Gaussian noise does not affect the performance of the model, thus confirming that only an AI-driven attack can suitably exploit this scenario.

As a final remark on these first results, we observe that our attack proved to be resistant to all the different countermeasures we considered.
In fact, as expected, the use of Local Differential Privacy gives boundaries to the adversary, allowing for a controlled impact of the attack on gradients, thus keeping them quite similar to benign ones and, therefore, very complex to detect.
The underlying assumption of the aforementioned defenses is that such a limited impact on the gradients, in principle, would completely prevent the effectiveness of the attack. However, the additional community-based privacy solution of the attacked scenario provides an opportunity to boost this malicious signal.

\begin{figure}
    \centering
    \includegraphics[width=\columnwidth]{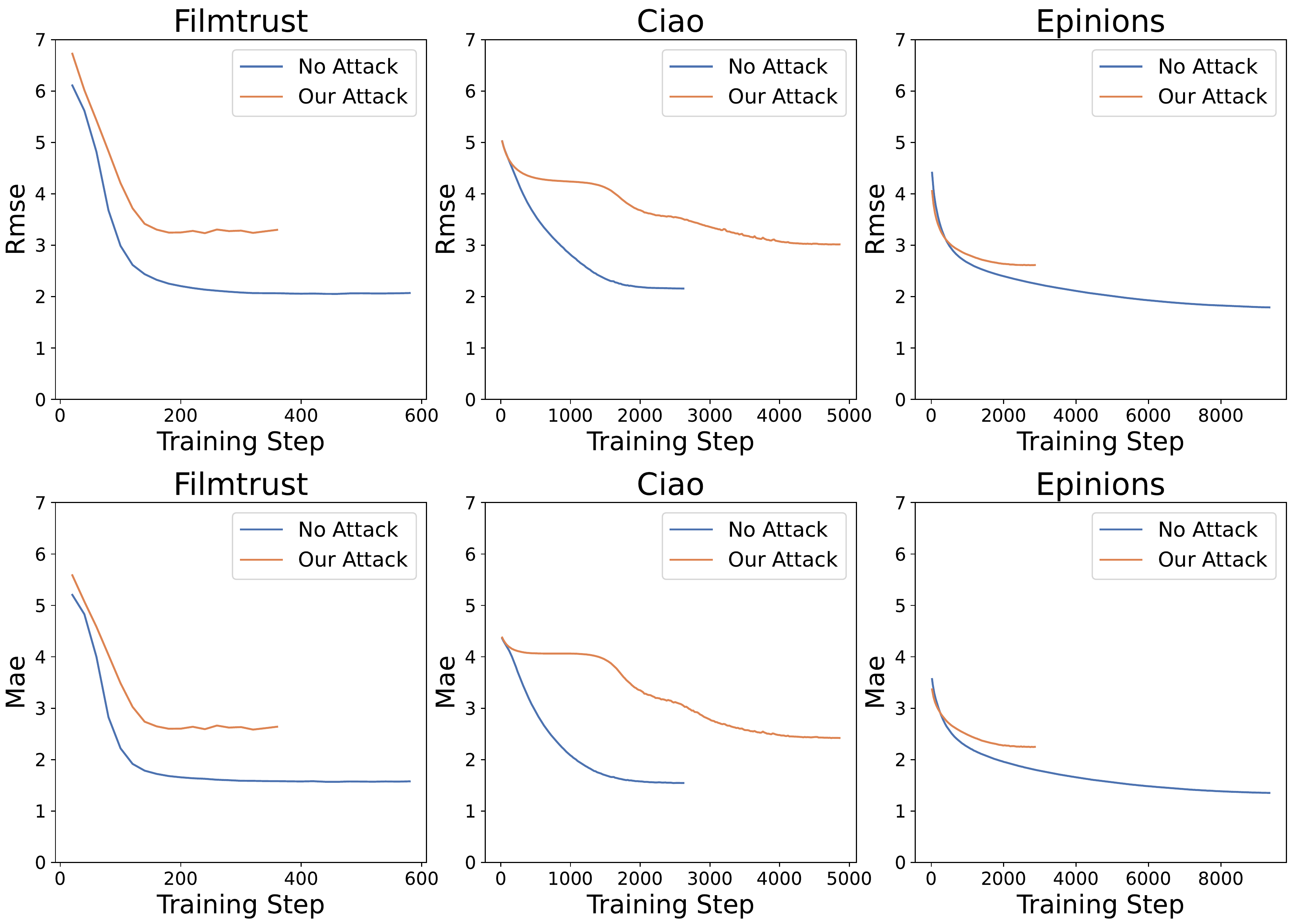}
    \caption{Performance of the federated learning model on the validation set with and without our attack. 
    \label{fig:Att-FeSoG}}
\end{figure}

To have a confirmation of our intuition, in Figure~\ref{fig:Att-FeSoG}, we show the variation of the performance metrics of the GNN model during the training phase. 
We can see at the very beginning of the training phase, the performances of the federated model on the validation set, with and without attacks, are almost identical. As the training continues, the difference between the normal and the attacked model increases, reaching high values by the end of it.
Indeed, after the first epoch, the clients surrounding the nodes controlled by the attacker begin to sample the malicious pseudo-items forged by them, thus permanently poisoning their local models. Such a mechanism continues, epoch by epoch, expanding the malicious signal to a growing neighborhood. In the end, all the poisoned clients will contribute to the attack boosting the negligible original signal produced by the attacker.

To deepen the analysis of this aspect, we tested our solution with both different percentages of malicious clients and several configurations of the number of pseudo-items sampled by the clients. 

\begin{figure}
    \centering
    \includegraphics[width=\columnwidth]{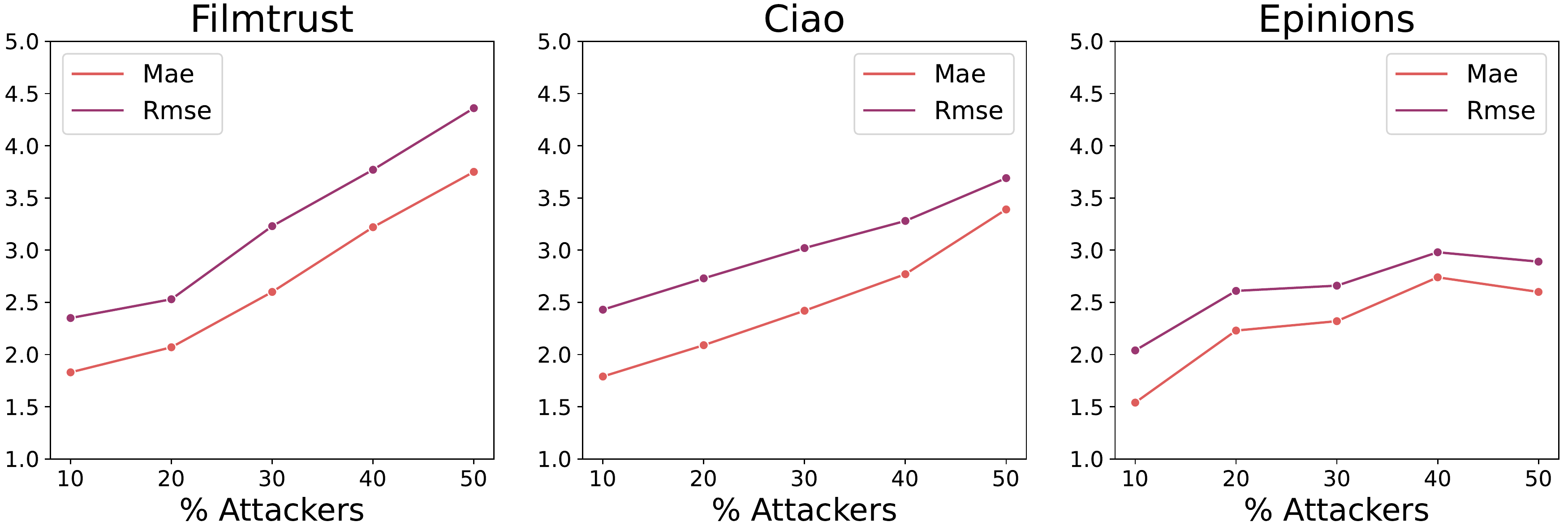}
    \caption{Performance of the federated learning model with different percentages of malicious clients.}
    \label{fig:PercAtt}
\end{figure}

In particular, in Figure~\ref{fig:PercAtt}, we show the impact of our attack on the performance of the federated learning GNN model underlying the {\em Main Scenario} with a percentage of clients controlled by the attacker spanning from $10\%$ to $50\%$.

\begin{figure}
    \centering
    \includegraphics[width=\columnwidth]{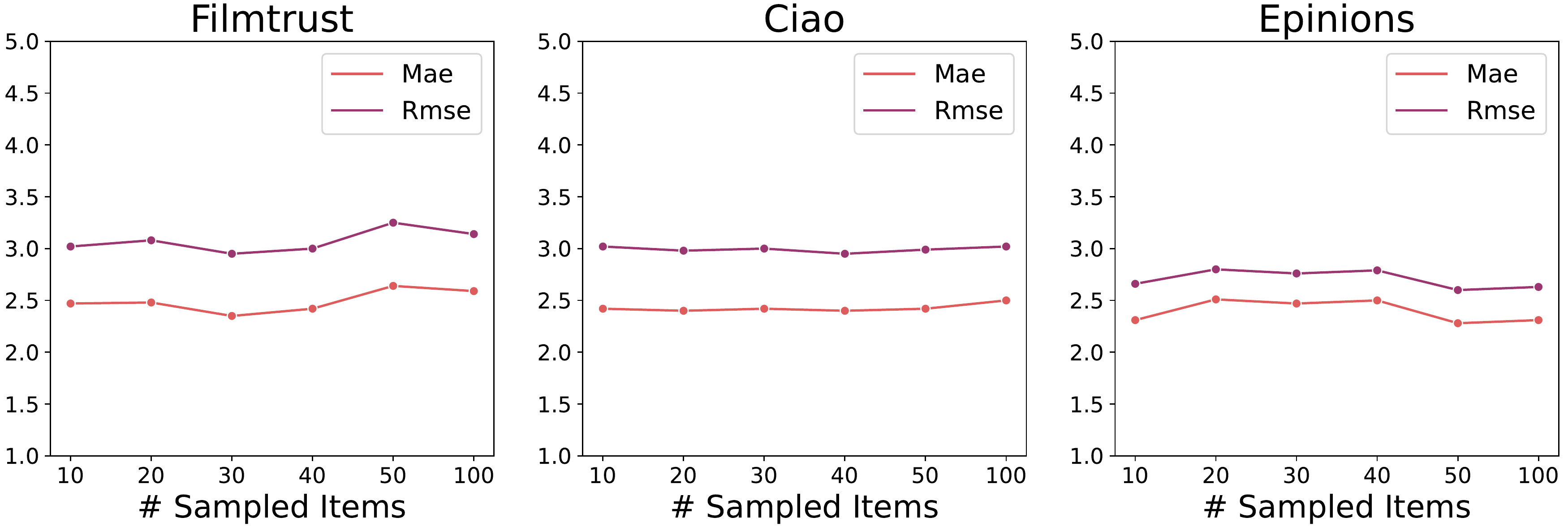}
    \caption{Performance of the federated learning model under our attack with different numbers of sampled pseudo-items per client.}
    \label{fig:sampledItems}
\end{figure}

As expected, from this figure, we can see that the increasing number of malicious clients causes a linearly related detriment to the model performance. However, the variation is not very steep and, sometimes, almost stable, proving that the attack strength does not depend only on the number of controlled malicious clients.
In Figure~\ref{fig:sampledItems}, we show the variation of the model performance for different configurations of the number of sampled pseudo-items (i.e., $\{10, \ 20, \ 30, \ 40, \ 50, \ 100\}$).
Here, we can see how changing the number of sampled items does not significantly affect the performance of the attack.
This indicates that, at least for the considered datasets, a small number of pseudo-items is enough to spread the malicious payload to a sufficiently large set of clients, which, then, will unknowingly act as additional collaborators of the attacker.

\subsection{Results: Backdoor Mode}
\label{sub:ExpBackdoor}

This section is devoted to presenting the results of the experiments carried out to validate the performance of the Backdoor Mode of our attack (see Section~\ref{sub:Backdoor} for details).

In this experiment, we randomly selected a target user $u_t$ from the set of users of each of our datasets and randomly sampled groups of $10$ items from the whole item pool, excluding those already belonging to the local graph of $u_t$.
At this point, we carried out our attack on the referring scenario to force the system to learn a backdoor for this set of items so that, for the only user $u_t$, the ratings associated with these items are controlled by the attacker.
In the scenario, we included again the state-of-the-art defense mechanisms for federated learning presented in Section~\ref{sub:ExpConvergence}.
To measure the effectiveness of our backdoor attack on this setting, we used the $FCR$ metric defined in Section~\ref{sub:Setup}.
This metric estimates how close the ratings of the selected items predicted for the target user are with respect to the values proposed by the attacker.
To return a reliable estimation, it also considers the general error of the regressor (standard error of estimate) to purge the evaluation from possibly wrong predictions related to the accuracy of the model.
The results of this experiment are reported in Table~\ref{table:ResBackdoor}. 

As visible in this table, our attack is capable of achieving an average $FCR$ score higher than $80\%$ with a maximum of $100\%$ against all the defenses. Another important result is, as expected that the performance (assessed with RMSE and MAE metrics) of the model on benign items is preserved for all three datasets. 

To have a ground truth to compare the obtained results with, we also measured the $FCR$ score in the case of no attack to exclude any success case related to the data distribution and not to the attack effect.
As we can see from the results, the maximum $FCR$ value in the absence of an attack (implying a situation in which, by chance, the real ratings are in-line with the attacker selection) is around $20\%$ on average, thus showing, once more, the effectiveness of our attack.

\begin{table*}[ht]
\centering
\scriptsize
\caption{Results of the deceptive rating injection attack}
\label{table:ResBackdoor}
\begin{tabular}{|c|c|ccc|ccc|ccc|}
\hline
\multirow{2}{*}{Attack} & \multirow{2}{*}{Defense} & \multicolumn{3}{c|}{Filmtrust}                             & \multicolumn{3}{c|}{Ciao} & \multicolumn{3}{c|}{Epinions}\\ \cline{3-11} 
              &    & \multicolumn{1}{c|}{RMSE}       & \multicolumn{1}{c|}{MAE}   & FCR & \multicolumn{1}{c|}{RMSE}       & \multicolumn{1}{c|}{MAE}   & FCR & \multicolumn{1}{c|}{RMSE}       & \multicolumn{1}{c|}{MAE}   & FCR\\ \hline
                  
No Attack & None    & \multicolumn{1}{c|}{2.06} & \multicolumn{1}{c|}{1.56} & 20\%  & \multicolumn{1}{c|}{2.16} & \multicolumn{1}{c|}{1.56} & 20\% & \multicolumn{1}{c|}{1.79} & \multicolumn{1}{c|}{1.36} & 30\%\\ \hline

Our Attack (Backdoor Mode) & FoolsGold      & \multicolumn{1}{c|}{2.07} &  \multicolumn{1}{c|}{1.55} & 80\%  & \multicolumn{1}{c|}{2.19}  & \multicolumn{1}{c|}{1.56} & 100\% & \multicolumn{1}{c|}{1.78}  & \multicolumn{1}{c|}{1.34} & 100\%\\ \hline

Our Attack (Backdoor Mode) & Flame      & \multicolumn{1}{c|}{2.05} &  \multicolumn{1}{c|}{1.57} & 80\%  & \multicolumn{1}{c|}{2.18}  & \multicolumn{1}{c|}{1.55} & 100\% & \multicolumn{1}{c|}{1.79}  & \multicolumn{1}{c|}{1.39} & 100\%\\ \hline

Our Attack (Backdoor Mode) & Krum      & \multicolumn{1}{c|}{2.03} &  \multicolumn{1}{c|}{1.54} & 80\%  & \multicolumn{1}{c|}{2.15}  & \multicolumn{1}{c|}{1.54} & 100\% & \multicolumn{1}{c|}{1.79}  & \multicolumn{1}{c|}{1.34} & 100\%\\ \hline

Our Attack (Backdoor Mode) & Trimmed Mean      & \multicolumn{1}{c|}{2.05} &  \multicolumn{1}{c|}{1.56} & 80\%   & \multicolumn{1}{c|}{2.19}          & \multicolumn{1}{c|}{1.56} & 100\% & \multicolumn{1}{c|}{1.79}          & \multicolumn{1}{c|}{1.34} & 100\%\\ \hline

\end{tabular}
\end{table*}

\subsection{Evaluation on a Real Recommender System }

As a final experiment, we proceed by testing our Backdoor Mode attack against a real-life recommendation system.
To do so, first of all, we designed a recommender system on top of the GNN-based model described in the previous sections. Such a model includes the embedding of users and items according to their interactions, which are described in the datasets of reference in this paper (see Section~\ref{sub:Setup}). Moreover, as stated in Section~\ref{sub:FeSoG}, given the embeddings of a user and an item, an estimate of the rating that the given user would assign to the target item can be obtained through the dot-product between their embeddings.
With this information, it is possible to build a recommender system capable of suggesting an item to a user if the estimated rating, according to the strategy above, is higher than a recommendation threshold $\delta$.
A possible strategy to set a value for $\delta$ could be to consider that, usually, an item is recommendable to a user if its estimated rating is close to the upper bound of the rating range (i.e., it is higher than the median value of the range). As such, $\delta$ should be a value equal to a fraction of the rating score range (e.g., for a maximum rating score equal to $10$, $\delta=0.5$ indicates that the recommendable items must have a rating score higher than half of the maximum rating score, that is a rating higher than $5$).

The objective of this experiment is to demonstrate that our backdoor attack can force a recommender system to suggest to a target user any item (also those that would normally receive a minimum rating score).
Of course, it can even be used in the opposite direction, that is, to force the removal of a good item from the set of recommendable ones for a target user.

\begin{figure}[ht]
    \centering
    \includegraphics[width=\columnwidth]{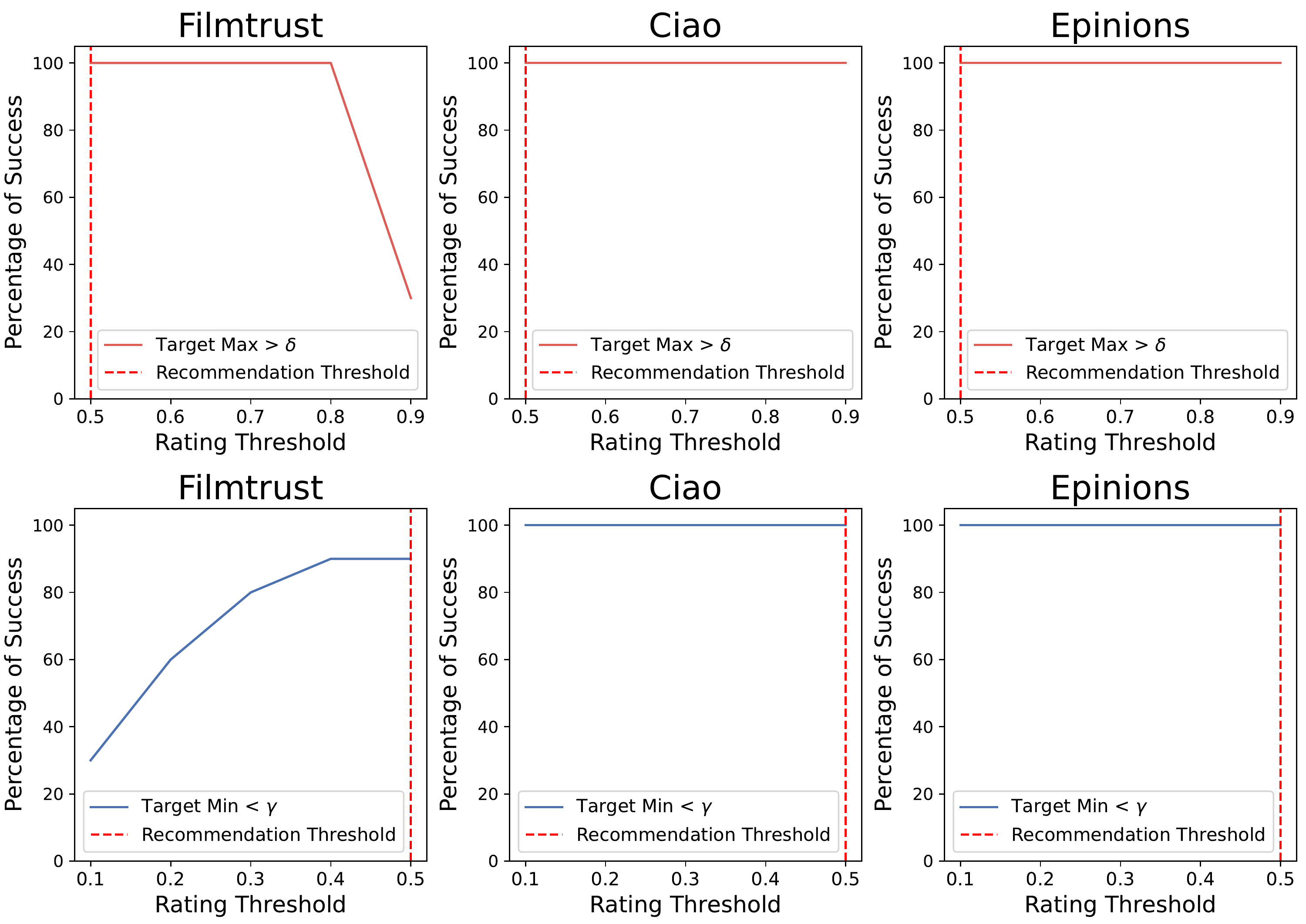}
    \caption{The recommendation system recommends an item if the rating is higher than the given threshold $\delta$. As a first possible objective, the attacker tries to force the model to predict an item of minimum rating as an item of maximum rating. We have a success when the rating exceeds a recommendation threshold $\delta$. Ex. $\delta=0.5$: rating $>$ $\delta \cdot max\_rating$.
    As a second objective, the attacker tries to remove a good item from the list of recommendable items. The goal is hence to reduce the rating for a target item under a negative threshold $\gamma \le \delta$. Therefore, we count the percentage of items with ratings lower than $\gamma$. Ex. $\gamma=0.4$: rating $<$ $\gamma \cdot max\_rating$.
    Worst case scenario: change the rating score of an item from a minimum to a maximum value and vice versa. 
    \label{fig:RecSys}}
\end{figure}

To properly configure our test, we started by selecting a target user and training the model in a safe configuration without attacks. Then, using the trained model, we estimated the rating of all the items available in relation to the target user.
After this, we sorted them and created a ranking of items for the target user.
As stated above, the goal of the attacker can be either to force the recommendation of a specific item to a target user or to remove a good item from the user recommendation list.
In both cases, we considered the worst-case situation, in which the specific item has originally an extremely low rating for the former objective, or an extremely high one, for the latter.
To obtain this configuration, as for the former objective, we selected the bottom $10$ items of the ranking above, and for the latter attack objective, we selected the top $10$ items as targets.
At this point, in our experiment, we tested the effectiveness of our Backdoor Mode attack against the above-introduced recommender system with different values of the recommendation threshold. In particular, to measure the obtained attack performances, we started with the former objective and counted the percentage of attacked items whose rating was higher than the recommendation threshold $\delta$. In this case, we considered different values of $\delta$, namely $\{0.5, \ 0.6, \ 0.7, \ 0.8, \ 0.9\}$, implying ratings for the recommendable items always above the median of the rating range and up to a value really close to the upper bound (i.e., $\delta=0.9$).
As for the latter objective, instead, we defined an additional negative threshold, called $\gamma$, to evaluate the attack strength.
The objective of this second threshold is the exact opposite of $\delta$, that is, to verify the percentage of items that have a lower rating than this negative threshold. Of course, the lower the negative threshold, the more complex the attack goal.
Also in this case, $\gamma \le \delta$ is obtained as a fraction of the maximum possible rating; in particular, we set it to $\{0.1, \ 0.2, \ 0.3, \ 0.4, \ 0.5\}$, respectively.
We reported the obtained results in Figure~\ref{fig:RecSys}. 

The first row of this figure shows the attack performance for the first objective, whereas the second row concerns the performance obtained for the second attack objective.
By analyzing this figure, we can see that, for both the Ciao and Epinions datasets, our attack is successful with all the possible threshold configurations for both objectives above.
As for the Filmtrust dataset, we can notice how the performance of our attack degrades to $30\%$ in the edge cases (i.e., the cases in which $\delta$ is equal to $0.9$, for the first objective, and $\gamma$ is $0.1$, for the second objective) while preserving its full effectiveness for the other configurations of the thresholds.
This behavior could be because this dataset has a fewer number of items concerning the others, thus increasing the probability for a single item to be sampled by multiple clients.     
In this way, the contribution of the attack could be partially overwritten by the benign clients' updates, which implies a slight reduction of the attack performance. 

\section{Related Work}
\label{sec:RW}


Federated recommender systems are becoming popular due to regulation on data and privacy of the users, like GDPR in the European Union~\cite{liu2022federated, wu2021fedgnn}.
This solution allows social media platforms to build effective recommender systems useful to produce high-quality suggestions while preserving the privacy of the final user.
However, this kind of collaborative strategy might be affected by malicious users that take part in the training of the federated recommender system~\cite{hitaj2017deep, zhang2022pipattack, christakopoulou2019adversarial}.

Christakopoulou et al.~\cite{christakopoulou2019adversarial}, proposed to use a Generative Adversarial Network (GAN) that generates fake users to be injected during the federated training to control the top-$K$ recommendations produced by the target recommender system. The proposed solution is designed to preserve the main characteristics of the data, thus ensuring unnoticeable changes.
Generative Adversarial Networks not only can be used to attack the systems in an adversarial way, but they are also effective in stealing private information from other users.
An example of that has been proposed by Hitaj et al.~\cite{hitaj2017deep} in which the attacker runs the collaborative learning algorithm and reconstructs sensitive information stored on the victim’s device.
The attacker also influences the training process inducing the victim to disclose more detailed information.

The conventional poisoning attacks on recommender systems, known as shilling attacks~\cite{lam2004shilling}, are not targeted to a specific type of recommender system. Therefore, the performance that they can achieve is sub-optimal to an attack targeted at a specific recommender system. Fang et al.~\cite{fang2020influence} proposed a series of techniques that optimize the attack to be more effective and achieve better performances compared to general shilling attacks.
Wu et al.~\cite{wu2022fedattack} proposed another optimized attack on recommender systems. In this paper, the authors proposed to use globally hardest sampling as a poisoning technique.
In particular, they retrieve pseudo “hardest positive samples” that are farthest from user embeddings to replace the original positive samples. The obtained gradients significantly impact the model convergence while being difficult to be perceived as malicious updates from the server. 
Fang et al.~\cite{fang2018poisoning} presented a poisoning attack optimized for graph-based recommender systems, like the attack we are proposing.
More in detail, in this poisoning attack, the authors' goal is to deceive the graph-based recommender system making it promote a target item to as many users as possible injecting fake users that give fake ratings to a subset of the items.

The superior ability of graph neural networks to learn graph-structured data makes them ideal for recommender systems~\cite{qiu2020exploiting}.
Considering this, Nguyen et al.~\cite{nguyen2023poisoning} proposed an attack that leverages both the representations of items and users to learn an optimal attack on a surrogate model. The proposed framework, similar to the one described above, synthesizes new users and associated edges to be added into a heterogeneous graph between real users and items before feeding the poisoned graph as input for optimization.
Graph-based recommender systems are also vulnerable to optimized backdoor attacks such as the one proposed by Zheng et al.~\cite{zheng2023link}.
In particular, the authors designed a backdoor attack against link prediction that injects nodes and uses gradient information to generate optimized triggers building a relationship between any two nodes in the graph to construct a general attack.


The GNN models are also prone to attacks that aim at the privacy of the model and the data.
These vulnerabilities could be exploited to infer group properties that are defined over the distribution of nodes and links, as proposed by Wang et al.~\cite{wang2022group}.
In particular, the authors designed six different attacks considering a comprehensive taxonomy of the threat model with various types of adversary knowledge.
They analyzed the main factors that contribute to group property inference attacks’ success and they found that it is possible to infer the existence of a target property by using the correlation between the property feature and a label in the target model.
Duddu et al.~\cite{duddu2020quantifying} designed three different attacks: the first infers if a node was included in the training graph, the second recreates the target graph, and the third infers sensitive attributes of the graph.
Considering attacks against the model instead, Zhang et al.~\cite{zhang2022inference} proposed a property inference attack that aims to infer the basic properties of the graph given the graph embeddings.

\section{Conclusions and Future Work}
\label{sec:Conclusion}

In this paper, we described an AI-based attack against a scenario composed of a privacy-preserving social recommender system leveraging Graph Neural Networks and federated learning to produce item recommendations.
Our attack design starts by analyzing the security of recent approaches aiming at building such recommender systems, including Differential Privacy and community-based strategies to improve sensitive data protection in federated learning contexts.
As a matter of fact, although, by design, one of the main features of federated learning is privacy protection, researchers have shown that, by analyzing local model updates produced by federated clients, it is possible to infer sensitive information concerning the local datasets.
For this reason, recent studies have included additional privacy protection strategies to face the above-mentioned issue.
This is the case of recent investigations in the context of social recommender systems, in which federated learning and Graph Neural Networks are adopted to build a predictive model to estimate item ratings to be fed to an underlying recommendation engine.
In such a scenario, some authors have proposed combining Differential Privacy modules with novel privacy-preserving strategies based on the main characteristics of the underlying scenario.
Indeed, in the context of social recommender systems, user interactions play a crucial role; this additional information allows the identification of communities of users related to each other.
Leveraging these communities for each client, it is possible to augment the training of their local model with knowledge derived from the other community members, thus creating an additional separation between the local updates and the training-sensitive data.
However, our intuition is that, if properly exploited, these additional privacy-preserving mechanisms can be used to produce a very impacting model poisoning attack against federated learning.

In this paper, we demonstrated this concept by designing a novel AI-based model poisoning attack with two operating modes, namely: {\em Adversarial Mode} producing a convergence inhibition effect and {\em Backdoor Mode} creating a deceptive rating injection attack on the federated model.
We tested our solution against a target social recommender system proposed by~\cite{liu2022federated} in a federated learning scenario equipped with the most effective state-of-the-art defenses.
The experimental results have shown how our attack is effective in all the considered cases. Moreover, to further show the significance of our achievements, we built a real-life recommender system to demonstrate that, with our attack operating in Backdoor Mode, an adversary can fully control the recommendations produced for specific target users.

The proposal described in this paper must not be intended as conclusive. Indeed, to demonstrate the general validity of our method, we are planning to extend our investigation by adapting the proposed attack strategy to other possible scenarios.
Moreover, the vulnerability we discovered is based on the collaborative nature of some privacy-preserving approaches for federated learning. For this reason, we intend to work on designing possible extensions of existing defenses to cope with the identified flaw.
Finally, we made explicit reference to a horizontal federated learning scenario. In the future, we plan to extend our research to vertical federated learning. Of course, due to the specificities of this variant, a thorough investigation must be carried out to understand how our attack methodology can be adapted to it. 


\end{document}